\documentclass[sigconf]{acmart}

\AtBeginDocument{%
  \providecommand\BibTeX{{%
    \normalfont B\kern-0.5em{\scshape i\kern-0.25em b}\kern-0.8em\TeX}}}

\copyrightyear{2023}
\acmYear{2023}
\setcopyright{acmlicensed}\acmConference[WWW '23 Companion]{Companion Proceedings of the ACM Web Conference 2023}{April 30-May 4, 2023}{Austin, TX, USA}
\acmBooktitle{Companion Proceedings of the ACM Web Conference 2023 (WWW '23 Companion), April 30-May 4, 2023, Austin, TX, USA}
\acmPrice{15.00}
\acmDOI{10.1145/3543873.3587537}
\acmISBN{978-1-4503-9419-2/23/04}

\usepackage{multirow}

\begin{document}

\title{A Semantic Partitioning Method for Large-Scale Training of Knowledge Graph Embeddings}


\author{Yuhe Bai \\ Supervised by Hubert Naacke and Camelia Constantin}
\affiliation{%
  \institution{Sorbonne University}
  \city{Paris}
  \country{France}}
\email{yuhe.bai@sorbonne-universite.fr}

\renewcommand{\shortauthors}{Yuhe Bai}

\begin{abstract}

In recent years, knowledge graph embeddings have achieved great success. Many methods have been proposed and achieved state-of-the-art results in various tasks. However, most of the current methods present one or more of the following problems: (i) They only consider fact triplets, while ignoring the ontology information of knowledge graphs. (ii) The obtained embeddings do not contain much semantic information. Therefore, using these embeddings for semantic tasks is problematic. (iii) They do not enable large-scale training. In this paper, we propose a new algorithm that incorporates the ontology of knowledge graphs and partitions the knowledge graph based on classes to include more semantic information for parallel training of large-scale knowledge graph embeddings. Our preliminary results show that our algorithm performs well on several popular benchmarks.

\end{abstract}

\ccsdesc[500]{Information systems~Data mining}
\ccsdesc[500]{Computing methodologies~Knowledge representation and reasoning}
\ccsdesc[500]{Computing methodologies~Parallel computing methodologies}

\keywords{knowledge graphs, link prediction, semantic partitioning, knowledge graph embeddings, parallel training}


\maketitle

\section{Introduction}

A knowledge graph (KG) is a structured representation of real-world information, where nodes represent entities (such as people, places, or concepts) and edges represent relations between these entities. Relations are represented as fact triplets, in the form of $\langle head\,entity, relation, tail\,entity \rangle$, abbreviated as $\langle h,r,t \rangle$, indicating that two entities are connected by a specific type of relation, e.g., $\langle France, hasCapital, Paris \rangle$. A KG is a directed heterogeneous multi-graph. Numerous KG datasets have been proposed, such as Freebase\cite{freebase}, DBpedia\cite{dbpedia}, YAGO\cite{yago}, and Wikidata\cite{wikidata}. These massive KGs can contain up to millions of entities and billions of facts. By representing knowledge in this way, a KG can be applied in various real-world applications, such as question answering\cite{qa1} and recommender systems\cite{recom1}.

An important aspect of KG is knowledge graph embeddings (KGEs), which are low-dimensional vector representations of entities and relations, making them more amenable to mathematical operations while preserving the information inherent in the KG. These entity and relation embeddings can then be used for a variety of downstream tasks, such as link prediction\cite{simple}, entity alignment\cite{ea1}, and entity typing\cite{et1}. In recent years, various KG embedding models, such as TransE\cite{transe}, RESCAL\cite{rescal}, and RotatE\cite{rotate}, have been proposed and achieved promising performance on these tasks. 

Currently, techniques for performing knowledge graph embedding tasks are mainly based on observed facts in KG. These techniques involve representing entities and relations in a continuous vector space and defining a scoring function for each fact to evaluate its plausibility. The embeddings for entities and relations are obtained by maximizing the overall plausibility of the observed facts. Thus, throughout the process, the learned embeddings only need to be compatible within each individual fact, and may not be predictive enough for downstream tasks\cite{survey1}\cite{rule1}. 

The motivation of this thesis is to not only utilize the fact triplets in KG, but also to incorporate other ontology information. 
At the same time, we aim to achieve this through a partitioning method that enables parallel training, making the trained embeddings contain more semantic information and perform better in downstream tasks. The code will be available on \url{https://github.com/YuheBAI/sem-kge}.

The structure of this paper is as follows: Section 1 is a brief introduction. In Section 2, we present the problem. Section 3 is the state-of-the-art and a discussion of the disadvantages of current methods. Our methodology is presented in Section 4 and 5, and current results are presented in Section 6. Finally, Section 7 is the conclusion and future work.

\section{Problem}

\sloppy{
Knowledge graph embedding methods aim to learn low-dimensional vector representations of entities and relations in a knowledge graph, represented as $KG=(\mathcal{E}, \mathcal{R}, \mathcal{T})$, where $\mathcal{E}$ is the set of entities, $\mathcal{R}$ is the set of relations, and $\mathcal{T}$ is the set of fact triplets $\langle h, r, t\rangle$ representing the relation between the head entity and the tail entity. These representations can then be utilized for a variety of downstream tasks, such as link prediction\cite{simple} and entity typing\cite{et1} (also known as entity classification) in the category of KG completion tasks. Even the largest Knowledge bases suffer from incompleteness. For example, in Freebase\cite{freebase} which contains 86M entities and 338M triples, 71\% of people have no known place of birth, and 75\% have no known nationality\cite{dong2014knowledge}. Thus, KG completion tasks, such as identifying new facts to add to KGs through link prediction, is an important research direction. 
}

\textbf{Link Prediction (LP)} in a KG is the task of exploiting existing facts to infer missing facts. This involves guessing the correct entity for a fact triplet such as $\langle h, r, ?\rangle$ in the tail prediction or $\langle ?, r, t\rangle$ in the head prediction. This task involves predicting new relations between entities based on their embeddings, which can be used to predict existing but unknown links, but also to predict future links and establish new connections and discover new knowledge in KG.

\textbf{Entity Typing (ET)} refers to the task of identifying the type(s) of entities in KGs. An entity in a KG can belong to multiple classes, making entity typing a multi-label classification problem. The goal is to assign one or multiple class/type labels to each entity. It's important to note that class labels are not stored in the fact triplets of the knowledge base, but in separate ontology files.

The typical process for using knowledge graph embedding techniques for various downstream tasks involves the following three steps:
(i) Embedding training: First, prepare the data to be trained, which contains the $\langle h, r, t\rangle$ triplets, entities, and relations. Then, train a knowledge graph embedding (KGE) model.
(ii) Downstream task: The next step involves applying the generated embeddings to a specific downstream task, such as link prediction, entity typing, or question answering.
(iii) Evaluation: This step involves evaluating the model's performance on the downstream task using different metrics. Common evaluation metrics used in link prediction include Mean Reciprocal Rank (MRR) and Hits@k, which measure the model's ability to rank predicted links correctly against the true links. For entity typing tasks, the evaluation metrics include precision, recall, and F1-score.

\paragraph{Problem statement} 

We first address the problem of learning embeddings for very large knowledge graphs when it is not feasible to use the entire KG in the learning process.
The importance of this issue stems from the ever-growing size of KGs, which makes the learning cost prohibitively high for very large graphs. There is an urgent need to find a way to learn embeddings at a limited computing cost.
This is not a straightforward task, as simply taking a random sample of a large KG results in low performance for link prediction tasks.

For example, if our task is to predict the missing nationality of individuals, we may use only the part of the KG that contains entities of type $Person$ along with directly related entities in the learning process. However, this approach may lead to low-quality predictions for nationalities. On the other hand, we may select another part of the KG that does not include all the people but includes entities indirectly connected to a person, which yields better-quality predictions.

Secondly, most existing knowledge graph embedding models do not take into account ontology information and face scalability issues when dealing with very large datasets due to their inability to consider parallel and distributed training. We believe that subgraph selection and graph partitioning are two sub-problems of a broader issue concerning the expression of the semantic description of the set of triplets.

Therefore, we define the following two sub-problems:

\begin{enumerate}
\item 
Given a knowledge graph of size $s$, a link prediction task for entities of type $P$, and a computing budget expressed as a percentage $p$ of the graph that can be processed, we aim to find the subgraph of size $p*s$ that maximizes the quality of the prediction task. Our goal is to propose a method that identifies the "essential" part of a very large KG based on the target task.

\item
Next, we plan to leverage the proposed method to tackle the problem of large-scale distributed KG embedding training. Given a large KG and a desired number of partitions, and given a KGE learning method, we aim to determine the triplets that should be placed together in each partition so as to achieve the same quality as if the embedding was learned in a centralized setting. Recent work has shown that KG partitioning can have an impact on quality in a distributed setting. We plan to apply the proposed method to the case of large-scale distributed KG embedding training.
\end{enumerate}

Most existing knowledge graph embedding processes do not consider semantic information and only utilize fact triplets. The goal of this thesis is to overcome the limitations of current solutions by incorporating ontology information in addition to fact triplets into the large-scale training of knowledge graph embeddings.

\section{State of the Art}

Various knowledge graph embedding models have been proposed in recent years. We use TransE\cite{transe} as an example since its idea is very simple and intuitive. Despite its simplicity, TransE and its variants continue to achieve state-of-the-art results on a range of tasks to this day. TransE\cite{transe} is a knowledge graph embedding model proposed by Bordes et al. in 2013. The basic idea behind the TransE model is straightforward, it assumes that the embedding of the head entity $h$ plus the embedding of the relation $r$ is approximately equal to the embedding of the tail entity $t$, that is: $h+r \approx t$. This means that $t$ should be a nearest neighbor of $h+r$ when $\langle h, r, t\rangle$ holds true, while $h+r$ should be far away from $t$ for negative samples.

TransE has proven to be effective on various tasks and is considered one of the simplest yet most successful knowledge graph embedding models. However, it has several drawbacks, such as its inability to handle one-to-many and many-to-many relationships. This has led to the development of improved models such as TransH\cite{transh}, TransR\cite{transr}, and TransD\cite{transd}, which are all based on translational distance models.

The current state-of-the-art in knowledge graph embedding models can be classified into three main categories: tensor decomposition based models, geometric models, and deep learning models.

\sloppy{
\begin{enumerate}
\item
\textbf{Tensor decomposition based models:} DistMult\cite{distmult}, ComplEx\cite{complex}, SimplE\cite{simple}, and TuckER\cite{tucker}, etc.

\item
\textbf{Geometric models:} TransE and its variants\cite{transe}\cite{transh}\cite{transr}\cite{transd}, RotatE\cite{rotate}, and HAKE\cite{hake}, etc.

\item
\textbf{Deep learning models:} ConvE\cite{conve}, R-GCN\cite{rgcn}, and CompGCN\cite{compgcn}, etc.

\end{enumerate}
}

\begin{table}
  \caption{Size statistics for YAGO 4 Full, Wikipedia (W), and English Wikipedia (E)}
  \label{yago4}
  \begin{tabular}{lrrr}
    \toprule
    Triples & Yago Full & Yago W & Yago E\\
    \midrule
    Facts & 343M & 48M & 20M\\
    rdf:type & 70M & 16M & 5M\\
    Labels & 303M & 137M & 66M\\
    Descriptions & 1399M & 139M & 50M\\
    Avg. facts per entity & 5.1 & 3.2 & 4\\
    Classes & 10124 & 10124 & 10124\\
  \bottomrule
\end{tabular}
\end{table}

However, most state-of-the-art methods have the following limitations: first, they only consider facts triplets when training the entity and relation embeddings, ignoring the huge semantic information in the KG dataset. As Table~\ref{yago4} shows, taking YAGO4 as an example, there are many types of triplets besides fact triplets, such as labels, descriptions, classes, and types. Other information besides facts should also be taken into account during the training of knowledge graph embedding models. 

Second, a recent study\cite{doemb} shows that current knowledge graph embedding methods do not actually capture knowledge graph semantics. They use the pre-trained model of the existing knowledge graph embedding methods to classify entities of different class levels. The results of the F1 score indicate that these embeddings perform well for high-level classes such as \textsl{Person}, but not for lower-level classes such as \textsl{Scientist}, suggesting that existing knowledge graph embedding models do not actually capture knowledge graph semantics, and their use for other semantic tasks may be questionable.

Recent studies have presented novel knowledge graph embedding models that incorporate auxiliary information from the knowledge base, such as entity types, textual descriptions, logical rules, etc, as the survey\cite{survey1} mentioned. For example, the SSE model\cite{sem2} requires entities of the same type to stay close to each other in the embedding space. The TKRL model \cite{sem1} utilizes a type encoder model for entity projection matrices to capture the type hierarchy. Furthermore, Zhang et al.\cite{sem3} extended traditional embedding methods by incorporating the hierarchical structure of relation clusters, relations, and sub-relations. 
Recently, OWL2Vec*\cite{owl2vec} proposed a method to encode the semantics of a KG ontology while learning the KG embeddings. The authors showed that OWL2Vec* outperforms KGE models such as TransE.

However, as far as we know, these ontology-based approaches for KGE embedding have been trained in a centralized manner and have only been empirically evaluated on rather small databases. Leveraging on these models, on a big data environment with very large-scale knowledge graphs, is an open research direction.
Two issues arise : (i) handling large graphs that do not fit in a single machine require partitioning methods.
(ii) distributed learning tend to decrease the embedding quality and overcoming this drawback is not straightforward.

Recent studies have evaluated the state-of-the-art techniques for parallel training of large-scale knowledge graph embedding models\cite{parallel} and graph embedding models\cite{spark}. Their experiments indicate that the optimal choice of partitioning and negative sampling technique varies with the dataset. Different partitioning and negative sampling methods can result in varying efficiency and effectiveness of knowledge graph embedding models. 
We argue that ontology-based KG partitioning is a key point in designing a new efficient parallel solution for learning KGE embeddings.

\section{Proposed Approach}

We are proposing:

\begin{enumerate}
\item 
A subgraph selection method that allows training of knowledge graphs to be performed within a constraint on graph size, resulting in improved quality compared to random sampling by utilizing semantic information to select crucial entities.

\item A partitioning method that enables parallel training, resulting in embeddings with enhanced semantic information and improved performance on downstream tasks.

\end{enumerate}

To address the issues present in current knowledge graph embedding models, we propose a novel approach that incorporates ontology information from the knowledge base. 
We term this approach \textbf{ontology-based KG partitioning}.
Our initial approach partitions the knowledge graph based on the class of each entity, grouping the fact triplets according to their class information.

Our algorithm introduces the following novelties regarding other existing knowledge graph embedding models: First, it leverages entity type information without altering the scoring function of existing models for fact triplets. Instead, it partitions these fact triplets based on their class, making our method flexible and compatible with most existing models. Second, our method is well-suited for parallel training. Our method proposes a novel approach that incorporates more semantic information from the knowledge base, leading to embeddings with increased semantic information and improved performance on downstream tasks.

\section{Methodology}
Our first objective is to evaluate the impact of semantic partitioning based on semantic classes on the prediction quality.

\begin{figure}[h]
  \centering
  \includegraphics[width=\linewidth]{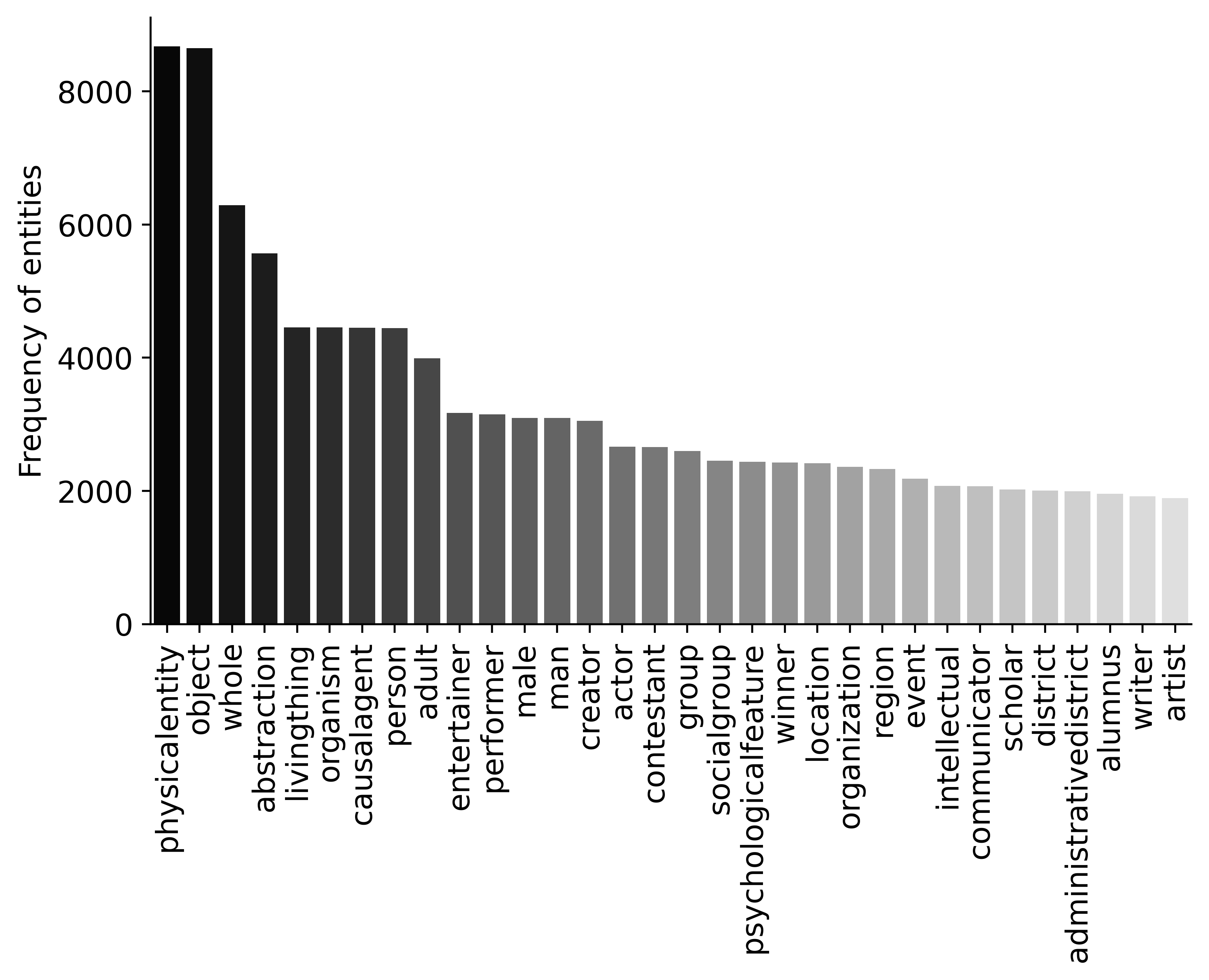}
  \caption{Freebase class frequency analysis.}
  \label{freebase_freq}
\end{figure}

As a first experiment, we utilized the facts, classes, and rdf:type information from the Freebase knowledge base. We then analyzed the frequency of each class. Fig~\ref{freebase_freq} illustrates the class frequency of Freebase\cite{freebase}.

\begin{table*}
\caption{Model performance in prior studies and our study (as percentages, on test data). First proposed: first reported performance on the respective datasets; Random partitioning: reproduced performance using random partitioning; Our semantic partitioning: performance with our method}
\label{result}
  
  \begin{tabular}{c c c c c c}
    \toprule
    & & \multicolumn{2}{c}{FB15K-237} & \multicolumn{2}{c}{FB15K} \\
    & Models & MRR & Hits@10 & MRR & Hits@10 \\
    \midrule
    \multirow{3}{*}{First Proposed} & TransE\cite{transe1}\cite{complex} & 29.4 & 46.5 & 38.0 & 64.1 \\
    & DistMult\cite{distmult1} & 24.1 & 41.9 & 65.4 & 82.4 \\
    & ComplEx\cite{distmult1} & 24.7 & 42.8 & 69.2 & 84.0 \\
    \midrule
    
    \multirow{3}{*}{Random Partitioning} & TransE & 29.0 & 48.5 & 69.2 & 85.9 \\
    & DistMult & 25.4 & 43.7 & 67.3 & 85.9 \\
    & ComplEx & 25.4 & 43.6 & 69.7 & 86.0 \\
    \midrule
    
    \multirow{3}{*}{Our Semantic Partitioning} & TransE & 29.1 & 48.6 & 69.2 & 85.9 \\
    & DistMult & 26.1 & 44.4 & \textbf{69.5} & \textbf{87.2} \\
    & ComplEx & 25.3 & 43.1 & 68.7 & 85.5 \\
    \bottomrule
  \end{tabular}
  
\end{table*}

As seen from Fig~\ref{freebase_freq}, the frequency of entities across different classes is imbalanced, with lower-level classes being less frequent than higher-level classes. We observe that including low-level classes is necessary to obtain an adequate number of partitions.
Furthermore, an entity can belong to multiple classes simultaneously. For instance, an entity can belong to both \textsl{Person} and \textsl{Writer}. 
Therefore, when partitioning, we assign each entity to its lowest class, which increases the number of partitions and allows for higher parallelism. We also believe that this should enhance the embedding quality because each partition contains more specific semantic information that can be reflected in the embeddings.

After assigning a class to each entity, in order to preserve the scoring function of existing models for fact triplets, we need to partition each fact triplet by assigning it to a specific class. This is done by classifying each fact triplet based on the class of its head entity. It's important to note that this is a preliminary, simple semantic partitioning method and requires further refinement based on the entity classification performance for different class levels.

To evaluate the results of our approach, two sets of evaluation metrics are conducted. The first one is the overall performance on link prediction tasks. For these tasks, we use commonly used evaluation metrics such as MRR (Mean Reciprocal Rank) and Hits@K. 
MRR is a measure of the average of the inverse of the rank of the first correct answer. Hits@K indicates whether at least one of the top-K recommended items is present in the test data.

The obtained embeddings are then utilized for entity classification tasks. For this purpose, the evaluation metrics that will be used are precision, recall, and F1 score. Precision measures the fraction of true positive predictions among all positive predictions made by the model. Recall measures the fraction of true positive predictions among all actual positive entities, and the F1-score is the harmonic mean of the precision and recall.

We then refine our semantic partitioning method based on the F1 scores obtained at different class levels. By fine-tuning our semantic partitioning method, our method can retain more semantic information in the obtained embeddings.

\section{Preliminary Results}

In this section, we present the current status of our work and the most significant results that have been reached so far.

\textbf{Datasets.}
In our study, we evaluated the performance of our proposed method on several benchmark datasets commonly used in the task of knowledge graph completion. FB15K\cite{transe} is a small subset extracted from Freebase, it contains 14,951 entities, 1,345 relations, and 592,213 triplets. However, many of these triples are inverses, which cause leakage from the training to testing and validation splits. FB15K-237\cite{fb15k-237} was created from FB15K by filtering out redundant and inverse relations to ensure that the testing and evaluation datasets do not have inverse relation test leakage. It comprises 310,079 triplets with 14,541 entities and 237 relations.

\textbf{Implementation.}
We implemented our semantic partitioning method and reproduced random partitioning for comparison, on top of the DGL-KE library\cite{dglke}, which provides a set of popular KGE models, training parameters, and evaluation techniques, and is capable of handling large-scale data. We considered three popular KGE models, TransE\cite{transe}, DistMult\cite{distmult}, and ComplEx\cite{complex}. For each model, we utilized the best hyperparameters proposed by DGL-KE for training and employed 64 CPUs on a single machine for parallel training. 

\textbf{Preliminary results.}
Preliminary results are shown in Table~\ref{result}. For each of the three models on each dataset, we compared its first proposed results, reproduced random partitioning results, and our semantic partitioning results. Our preliminary results show that the performance of our semantic partitioning method is model-dependent. For the TransE model, our method only slightly differs from random partitioning, for the ComplEx model, our method yields inferior results to random partitioning, but for the DistMult model, our method provides a significant improvement over random partitioning. This may be due to the scoring function of different models. We plan to conduct entity classification on the obtained embeddings at different levels to further optimize our method. It should be noted that even if some models are not as good as random partitioning in overall results, our model may perform better on low-level classes. Further evaluations are required to verify this hypothesis. If we look at the preliminary link prediction results we obtained now, our method performs well in overall.

\section{Conclusion and Future Work}

To conclude, to address the issues present in current knowledge graph embedding models, this paper presents a new semantic partitioning method for training knowledge graph embeddings in parallel. It leverages the ontology information in the knowledge base for partitioning to preserve more semantic information. It is very flexible and can be adapted to most existing knowledge graph embedding methods, and can also be used for parallel training. Our preliminary results show that our method performs well on several popular benchmarks. Future work will be divided into the following parts: (1) Conduct entity classification task and evaluate the F1 score on different class levels. (2) Fine-tune our method to make it more sophisticated and preserve more semantic information. (3) Evaluate the performance of our method on very large-scale datasets and compare its efficiency with other partitioning methods.

\begin{acks}
This work has been funded by the SCAI Sorbonne Center for Artificial Intelligence.

\end{acks}




\end{document}